\documentclass[letterpaper, 10 pt, conference]{ieeetrans}  

\IEEEoverridecommandlockouts                              

\usepackage[utf8]{inputenc}
\usepackage{mathtools}
\usepackage{amsmath}
\usepackage{amsfonts}
\usepackage{amssymb}
\usepackage{balance}
\usepackage{adjustbox}
\usepackage{lipsum}  
\usepackage{xcolor}
\usepackage{graphicx}
\usepackage{booktabs}
\usepackage{hyperref}
\usepackage[sorting = none, backend = bibtex, style=ieee]{biblatex}
\usepackage{svg}
\usepackage{tabularx} 
\usepackage{dblfloatfix}
\usepackage{array}
\usepackage{caption}
\usepackage{titleps}

\AtNextBibliography{\small}

\definecolor{somegray}{rgb}{0.5, 0.5, 0.5}
\newcommand{\darkgrayed}[1]{\textcolor{somegray}{#1}}
\makeatletter
\newcommand*\titleheader[1]{\gdef\@titleheader{#1}}
\AtBeginDocument{%
  \let\st@red@title\@title
  \def\@title{%
    \vskip-2em
    \bgroup\normalfont\large\centering\@titleheader\par\egroup
    \vskip1.5em\st@red@title}
}
\makeatother

\titleheader{\darkgrayed{This paper has been accepted for publication at the\\
IEEE International Conference on Robotics and Automation (ICRA)
, 2026.
\copyright IEEE}
}

\title{\LARGE \bf Global End-Effector Pose Control of an Underactuated Aerial Manipulator via Reinforcement Learning}

\author{Shlok Deshmukh$^{1}$, Javier Alonso-Mora$^{1}$, and Sihao Sun$^{1}$ 
\thanks{$^1$ Authors are with the Department of Cognitive Robotics (CoR), Faculty of Mechanical Engineering, Delft University of Technology, Delft, Netherlands}}

\begin{document}

\maketitle

\begin{abstract}
Aerial manipulators, which combine robotic arms with multi-rotor drones, face strict constraints on arm weight and mechanical complexity. In this work, we study a lightweight 2-degree-of-freedom (DoF) arm mounted on a quadrotor via a differential mechanism, capable of full six-DoF end-effector pose control. While the minimal design enables simplicity and reduced payload, it also introduces challenges such as underactuation and sensitivity to external disturbances. To address these, we employ reinforcement learning, training a Proximal Policy Optimization (PPO) agent in simulation to generate feedforward commands for quadrotor acceleration and body rates, along with joint angle targets. These commands are tracked by an incremental nonlinear dynamic inversion (INDI) attitude controller and a PID joint controller, respectively. Flight experiments demonstrate centimeter-level position accuracy and degree-level orientation precision, with robust performance under external force disturbances—including manipulation of heavy loads and pushing tasks. The results highlight the potential of learning-based control strategies for enabling contact-rich aerial manipulation using simple, lightweight platforms. Videos of the experiment and the method are summarized in \url{https://youtu.be/bWLTPqKcCOA}.

\end{abstract}
\section{Introduction}
Aerial manipulation combines the dexterity of robotic arms with the mobility of quadrotors, enabling in-flight tasks such as pushing, sliding, grasping, and transporting objects \cite{9462539,5513152,6094871}. 
These capabilities are valuable for applications including material transport, disaster response, industrial inspection and maintenance, where they can reduce cost and risk for human workers. 
However, deploying aerial manipulators in the real world is challenging due to the design constraints (space, weight, and energy), as well as control challenges due to nonlinear coupled dynamics between the flying base and the arm, together with disturbances arising from interaction with the environment. 

\begin{figure}
    \centering
    \includegraphics[width=1.0\linewidth]{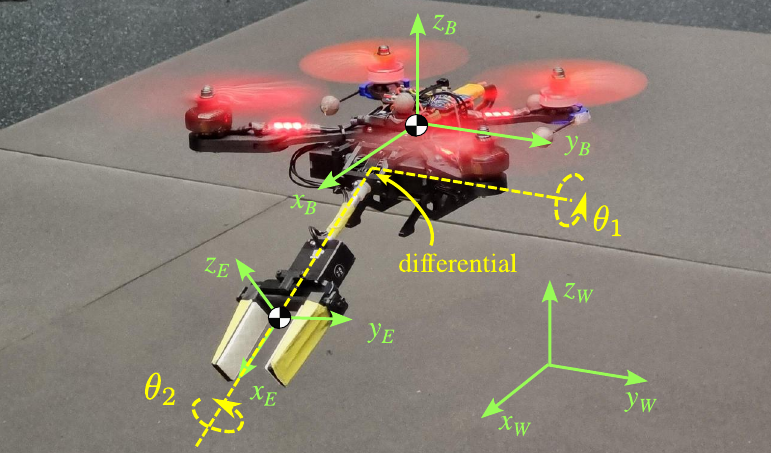}
    \caption{Snapshot of a DSAM aerial manipulator, consisting of a quadrotor base with an arm mounted through a differential mechanism that provides 2-DoF motion.}
    \label{fig:eye-catcher}
\end{figure}

Many aerial manipulators employ redundant arms with multiple degrees of freedom, or add additional tilted / tiltable rotors to achieve full actuation of the base, but at the cost of additional weight and structural complexity (e.g., ~\cite{kim2013aerial, tognon2019truly, afifi2022toward}).
Unlike ground and underwater robots, aerial robots must prioritize lightweight design due to their limited thrust-to-weight ratio and restricted onboard energy resources.
To address this, we adopt a minimal yet effective design: the \textit{Differential Shoulder Aerial Manipulator (DSAM)}, where a single rigid arm with an end-effector (e.g., a gripper) is connected to the quadrotor base through a differential driven by two rotational actuators, enabling pitch and roll motions~(Figure~\ref{fig:eye-catcher}).
Prior work has shown that a configuration for AM like DSAM is sufficient to achieve \textit{global} 6-DoF end-effector pose regulation, despite its mechanical simplicity~\cite{welde2021dynamically}.
However, the problem of designing robust whole-body control algorithms for DSAM remains largely unexplored.

Before developing model-based control methods, we first investigate reinforcement learning (RL) for the whole-body control of DSAM.
RL-based controllers offer significant advantages by alleviating the need for analytical models and their simplifications, implicitly considering the system actuation constraints through explorations, enabling fast inference, and supporting adaptation to unmodeled disturbances through domain randomization, all of which are essential for real-world aerial manipulation.
To address the sim-to-real gap, we also employ domain randomization during training, and adopt the scheme that hybridizes the RL with downstream low-level controllers~\cite{hwangbo2019learning}.
Specifically, we employ an incremental nonlinear dynamic inversion (INDI) attitude controller for the quadrotor base, and a PID controller for the arm.
We have experimentally validated the proposed algorithm in controlling the 6-DoF end-effector pose, as well as experiments testing the robustness against external forces on the end-effector, including carrying an additional load, and pushing an object.

Our contributions are summarized as follows:
\begin{enumerate}
    \item We develop and deploy a reinforcement learning policy, trained using Proximal Policy Optimization (PPO), for whole-body control of the Differential Shoulder Aerial Manipulator (DSAM). The policy achieves centimeter-level position accuracy and degree-level orientation precision in real-world experiments.
    \item We demonstrate robust end-effector pose control under significant external disturbances, including lifting a payload of 140~g (over 16\% of system mass) and pushing a 590~g object (over 68\% of system mass).
    \item Through ablation studies, we show that body-rate and joint observations, high-level control abstractions, and domain randomization of mass and friction significantly improve sim-to-real transfer, as evidenced by learning curve analysis.   
\end{enumerate}

\section{Related Work}
\subsection{System Design}
Aerial manipulators can be grouped into two design families, each aiming to realize end-effector (EE) regulation in SE(3) or a relevant subspace. (i) The first augments an underactuated unidirectional multirotor base with a serial or parallel arm (e.g., \cite{kim2013aerial,chen2025aerial}). (ii) The second employs overactuated drone bases with tilted or tiltable rotors, allowing EEs fixed to the base to realize arbitrary pose tracking in SE(3) (e.g., \cite{rashad2019port,li2024servo,kamel2018voliro,hui2024passive,tognon2018omnidirectional}). 
Some designs employ both choices (e.g., ~\cite{tognon2019truly, afifi2022toward, malczyk2023multi}).
A more detailed testimony of aerial manipulator designs, and their comparisons, is given in the survey~\cite{9462539}.
We focus on the first family and adopt the DSAM: a quadrotor base connected to an arm through a 2-DoF differential. 
DSAM is one of the most minimal designs achieving \textit{global} 6-DoF end-effector pose control, where the dual rotational actuators enlarge arm torques, thereby boosting load and manipulation capability without increasing weight or complexity. 
Ref.~\cite{welde2021dynamically} concludes that task flatness holds for a DSAM in a special condition, namely, the differential is located at the CoM of the quadrotor base.  
In practice, however, this condition may not hold due to design constraints and CoM variations. 
In addition, directly designing the control law through differential flatness cannot effectively satisfy input and state constraints, which are essential for aerial robotic manipulators.

\subsection{Planning and Control}
Specifically for DSAM, a controller still needs to be designed to reliably track setpoints and trajectories across the full SE(3) task space.
Most research on aerial manipulator control employ model-based methods. Reactive control methods, such as feedback linearization and geometric control \cite{lippiello2012exploiting, Zhang_2020,9345436, 6386021,6580711}, offer computational efficiency but provide limited mechanisms for handling actuator bounds and state constraints.  
Several works employ model predictive control (MPC) and Model Predictive Path Integral (MPPI) control for aerial manipulators, which improve constraint handling and can reason over short horizons, yet they incur substantial online computation and require careful contact and cost-shaping design in contact-rich tasks~\cite{7139850, marti2021full, 9830871}. 
Motivated by promising progress in robot learning for manipulation (e.g., \cite{rajeswaran2017learning}), recent studies have begun to explore reinforcement learning for aerial manipulators, demonstrating promising results on platforms with fully-actuated bases (e.g., omnidirectional bases with fixed EEs for door opening \cite{cuniato2023learning}). 
Building on this trend, we present the first whole-body controller for DSAM, based on learning-based method, that globally tracks SE(3) setpoints and addresses the underactuation of the DSAM while embracing its hardware design simplicity. 
This work will pave the way to future research for DSAM in contact-rich aerial manipulation tasks.

\section{Preliminaries}
\label{section:preliminary}
Three coordinate frames are considered in this work:
\begin{enumerate}
    \item World frame ${\{{x}_{W}, {y}_W, {z}_{W}}\}$ - inertial frame of reference.
    \item Body frame ${\{{x}_B, {y}_B, {z}_B}\}$ - located at the drone platform's CoM.
    \item End-effector frame $\{{{x}_{E}, {y}_{E}, {z}_{E}}\}$ - located at gripper's CoM.
\end{enumerate}
Note that for dynamic consistency, we define the end-effector frame at the CoM of the gripper, rather than at its tip.
Modeling each link and arm as a rigid body, we define the configuration and velocity of the system as
\begin{equation}
\boldsymbol{q} =
\begin{Bmatrix}
{}^{W}\!\boldsymbol{p}_{b} \\
\boldsymbol{R}_{wb} \\
\boldsymbol{\theta}
\end{Bmatrix}
\in \mathbb{R}^3 \times SO(3) \times \mathbb{T}^2,
\quad
\boldsymbol{v} = 
\begin{bmatrix}
{}^{W}\!\dot{\boldsymbol{p}_{b}} \\
{}^{B}\boldsymbol{\Omega}_{b} \\
\dot{\boldsymbol{\theta}}
\end{bmatrix}
\in \mathbb{R}^{8},
\end{equation}
where ${}^{W}\!\boldsymbol{p}_{b}$ is the position of the system in world frame, $\boldsymbol{R}_{wb}$ the rotation from world to body frame and $\boldsymbol{\theta}$ are joint angles. 

The system evolves according to coupled nonlinear dynamics given by generalized Lagrangian method presented in \cite{From2012SingularityFreePartTwo}
\begin{equation}
\boldsymbol{M}(\boldsymbol{q}) \dot{\boldsymbol{v}} + \boldsymbol{C}(\boldsymbol{q}, \boldsymbol{v}) \boldsymbol{v} + \boldsymbol{G}(\boldsymbol{q}) =
\boldsymbol{\tau}(\boldsymbol{q}, \boldsymbol{u}),
\label{eq:full_dynamics}
\end{equation}

where $\boldsymbol{\tau}$ represents the generalized forces due to control inputs $\boldsymbol{u}:= [\boldsymbol{\omega}_{cmd},~\boldsymbol{\tau}_{\theta}]^\top$ and configuration $\boldsymbol{q}$. Here, $\boldsymbol{\omega}_{cmd} \in \mathbb{R}^{4}_{\geq0}$ denotes rotor speed commands for the quadrotor base, and $\boldsymbol{\tau}_\theta \in \mathbb{R}^2$ denotes joint torques.

\begin{figure*}[t!]
    \includegraphics[width=1.0\textwidth]{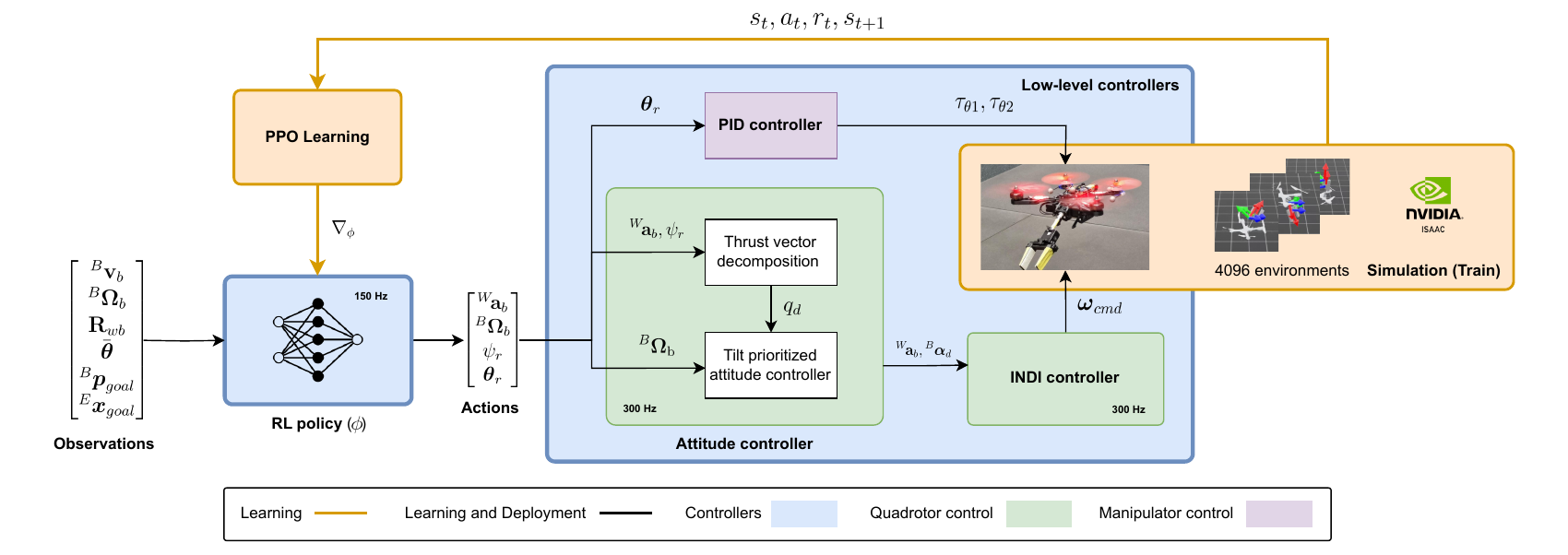}
    \captionof{figure}{Overview of proposed controller architecture and training methodology.}
    \label{fig:control_architecture}
\end{figure*}

Our objective is to control the end-effector pose of the system, defined by the forward kinematics:

\begin{equation}
         {}^{W}\!\boldsymbol{x}_{e} := \left[ {}^{W}\!\boldsymbol{p}_e,~ \boldsymbol{q}_{we}\right]^\top =  f_f(\boldsymbol{q})  \in SE(3)
\end{equation}
where ${}^{W}\!\boldsymbol{p}_e$ is end-effector position in world frame and $\boldsymbol{q}_{we}$ is the quaternion from the world to end-effector frame.

We formulate the control task learning a policy at each time $t$, maps the observed state $s_t \in \mathcal{S}$ to an action $a_t \in \mathcal{A} \subseteq  \mathbb{R}^9$,
which is then mapped to valid control inputs $\boldsymbol{u}_t \in \mathbb{R}^{6}$ using low-level controllers ($LLC$):
\begin{equation}
\pi_\phi: \mathcal{S} \rightarrow \mathcal{A}, \qquad 
a_t = \pi_\phi(s_t), \qquad
\boldsymbol{u}_t = LLC(a_t)
\end{equation}
During learning, at time $t$ the task is scored by an instantaneous reward:
\begin{equation}
r_t \;=\; r(s_t, a_t) 
\end{equation}

The control inputs \( \boldsymbol{u}_t \) directly influence the system dynamics through (\ref{eq:full_dynamics}).
The rewards aim to minimize the tracking error between the actual and desired end-effector poses \( {}^{W}\boldsymbol{x}_{e}(t) \) and \( {}^{W}\boldsymbol{x}_{goal}(t) \), in the presence of unknown external wrench acting on the end effector.

\section{Methodology}
\subsection{Control architecture}
\label{section:control_architecture}
The control architecture is presented in Fig.~\ref{fig:control_architecture}, which integrates the trained policy with low-level controllers. 
We employ a hierarchical structure that uses an RL policy as an outer-loop controller, which generates linear acceleration, body rate,  heading angle, and joint angle commands for the inner-loop controllers to follow.
The inner-loop controllers include a PID controller for the differential joints, and an INDI-based controller for the quadrotor base, to address modeling uncertainties and external wrench disturbances.

\subsubsection{Reinforcement Learning Policy}
The RL-policy is a multi-layer Perceptron (MLP) with three hidden layers of sizes 512, 256, 128 neurons respectively, and ELU activation functions.

It receives observations from the environment and generates actions for subsequent controllers at 150 Hz.
Mean and Variance statistics from training are used to normalize the observation vector, followed by inference using ONNX Runtime engine \cite{onnxruntime} onboard for shorter inference times. 

The RL-policy's \textbf{action space} includes desire linear accelerations, body rates, and yaw angles of the quadrotor base (${}^{W}\!\boldsymbol{a}_{b}, {}^{B}\boldsymbol{\Omega}_{b}, \psi_r$), as well as the normalized desired joint angle of the arms $\bar{\boldsymbol{\theta}}_r \in [-2,~2]$, which is then scaled to $\boldsymbol{\theta}_r \in  [-\pi/2,~\pi/2]$.

The \textbf{observation space} of the RL-policy at any time step is a $29$-dimensional vector that includes the following variables: Linear velocity of the quadrotor base in the body frame ${}^{B}\dot{\mathbf{x}}_b$; body rate of the quadrotor base ${}^{B}{\boldsymbol{\Omega}}_{b}$; rotation matrix of the quadrotor base ${\boldsymbol{R}}_{{wb}}$; normalized joint angles $\bar{\boldsymbol{\theta}}\in[-1,~1]$; goal position in the quadrotor frame ${}^{B}\boldsymbol{p}_{\text{goal}}$; end-effector pose error, expressed as the goal pose in the end-effector frame, ${}^{E}\boldsymbol{x}_{\text{goal}}$. 

\subsubsection{Inner-loop Quadrotor Attitude Controller}
The inner-loop attitude controller, running at 300 Hz, converts the RL outputs into rotor-speed commands $\omega_\mathrm{cmd}$ for the quadrotor base. 
It first maps the RL-generated acceleration and heading to a desired attitude via thrust vector decomposition, and then combines a tilt-prioritized controller~\cite{brescianini2018tilt} with incremental nonlinear dynamic inversion (INDI)\cite{sun2022comparative} to compute rotor speeds, using the RL-provided body rates as feedforward terms. 
This inner-loop attitude controller has demonstrated robustness to model mismatches in the quadrotor’s rotational dynamics, which is particularly important for DSAM, where arm motion induces torque disturbances on the base. For further details, we refer readers to~\cite{sun2022comparative}.

\subsection{Training of the RL Policy}
\label{section:training} 

We model the control architecture (Fig.~\ref{fig:control_architecture}) in our training environment and PPO~\cite{schulman2017proximal} to train our agent to perform end-effector pose control. 

\subsubsection{Training Environment}
\label{subsec:training_env}

We use Isaac Lab~\cite{mittal2023orbit} for training the RL-policy as it extensively utilizes GPUs for both simulation and learning, enabling high data throughput and faster learning. The policy training is implemented using SKRL reinforcement learning library~\cite{serrano2023skrl}, which provides an integration with Isaac Lab. We model the DSAM system in the training environment. 
Apart from modeling the control architecture (Section \ref{section:control_architecture}) with low-level controllers, we also model filters and first-order dynamics of rotors.
 
PPO offers control over multiple aspects of the algorithm, making hyperparameter selection challenging. We begin with the hyperparameters used in \cite{rudin2022learningwalkminutesusing} as a starting point.

We train on 4096 environments in parallel, with a rollout ($n_{steps}$) size of 24 {(Batch size$= 4096 \times 24$)} and environment step frequency of 150 Hz. This corresponds to simulating 0.16 seconds of real time per environment, every training iteration. Our maximum episode length is 6 seconds, after which the environment gets timed out and reset, meaning a single episode will have multiple policy updates.

 To achieve a working policy, we train for 2 billion environment steps ($4096 \times 500,000$) that take 5.5 hours on a consumer-grade GPU (NVIDIA RTX 2080 Ti). 

\subsubsection{Rewards}
We define two primary rewards, one for position and one for orientation, along with additional penalty terms to smooth the action space and penalize large action magnitudes.  
All rewards are scaled by a negative exponential transformation that maps raw reward values between $(0, 1]$, enabling comparable weighing of individual reward components.

\textit{Position reward:} We use the L2 norm of the difference between the end-effector and goal positions as the basis for position reward:
\begin{equation}
{r}_{\text{pos}} = w_1 \cdot \textrm{exp}(-\alpha_{1} \cdot \| {}^{W}\!\boldsymbol{p}_{e} - {}^{W}\!\boldsymbol{p}_{goal} \|_2).
\end{equation}
The $\alpha_1$ value tunes reward sensitivity for local convergence. Larger values increase the gradient magnitude near the target but can lead to vanishing gradients far from the target, as the exponential saturates near zero.

\textit{Orientation reward}: We use the smallest geodesic angular difference between the orientations of the end-effector and the goal $\mathcal{E}_{\text{ori}} = 2 \arccos \left( \left|\boldsymbol{q}_{e}\cdot \boldsymbol{q}_{goal}\right| \right)$ to represent the orientation error of the end-effector.
We do not penalize the quadrotor's orientation directly, as the end-effector orientation already constrains the possible quadrotor orientations:
\begin{equation}
{r}_{\text{ori}} = w_2 \cdot \textrm{exp}(-\alpha_{2} \cdot \mathcal{E}_{ori}).
\end{equation}

\textit{Quadrotor action smoothing reward}: The norm of the difference between the quadrotor base action values $
{}^{\text{base}}\boldsymbol{a} = \left\{{}^{W}\!\boldsymbol{a}_b,\, {}^{B}\boldsymbol{\Omega}_b,\, {}^{W}\!\psi_{r}\right\}$
between the current and previous time steps penalizes large deviations in consecutive actions:
\begin{equation}
{r}_{\text{ds}} = 
w_3 \cdot \textrm{exp}(-\alpha_{3} \cdot 
\left\| 
{}^{\text{base}}\!\boldsymbol{a}_{t-1} - {}^{\text{base}}\!\boldsymbol{a}_t
\right\|_2^2).
\end{equation}
We found this term beneficial for reliable deployment.

\textit{Joint action smoothing reward:} The absolute difference between joint position action commands (${}^{\text{joint}}\boldsymbol{a} = \boldsymbol{\theta}_r$) is penalized to encourage smoother joint actions, producing near-constant torque with fewer jerks that might destabilize the platform:
\begin{equation}
{r}_{\text{js}} =
w_4 \cdot \textrm{exp}(-\alpha_{4} \cdot 
\left\|
{}^{\text{joint}}\!\boldsymbol{a}_{t-1} - {}^{\text{joint}}\!\boldsymbol{a}_t
\right\|_1).
\end{equation}

\textit{Quadrotor action magnitude reward:} The magnitude of quadrotor base actions ${}^{\text{base}}\boldsymbol{a} = \left\{{}^{W}\!\boldsymbol{a}_b,\, {}^{B}\boldsymbol{\Omega}_b,\, {}^{W}\!\psi_{r}\right\}$ is penalized to minimize oscillations and overshoot while reaching the goal pose:
\begin{equation}
{r}_{\text{dmag}} = w_5 \cdot \textrm{exp}(-\alpha_5 \cdot \left\| {}^{\text{base}}\boldsymbol{a}_t \right\|_2^2).
\end{equation}

The total reward is the weighted sum of all components:
\begin{equation}
\mathcal{R}_t = {r}_{\text{pos}} + {r}_{\text{ori}} + {r}_{\text{ds}} + {r}_{\text{js}} + {r}_{\text{dmag}}.
\end{equation}

\begin{table}[h]
\centering
\begin{tabular}{ccc}
\toprule
\textbf{Index} & \textbf{Weight} \( w_i \) & \textbf{Scale} \( \alpha_i \) \\
\midrule
1 & 4.0 & 1.2 \\
2 & 1.0 & 1.0 \\
3 & 0.5 & 1.0 \\
4 & 1.0 & 1.0 \\
5 & 0.1 & 1.0 \\
\bottomrule
\end{tabular}
\caption{Reward weights \( w_i \) and scaling factors \( \alpha_i \) used in the total reward function.}
\end{table}

\subsubsection{References and Resets}

We sample goal pose references\footnote{We use the terms \emph{references} and \emph{commands} interchangeably, following Isaac Lab terminology for references in goal-conditioned tasks} for the end-effector (gripper) during training. A new goal pose is sampled upon environment resets, which occur when an episode terminates either due to a timeout or when the quadrotor falls below 0.3 m.  

Positions are sampled from the ranges $(-1.0, 1.0)$~meter in the $X$–$Y$ directions and $(3.0, 5.0)$~m in the $Z$ direction. Orientations are sampled from $(-180^\circ, +180^\circ)$ in yaw and $(-90^\circ, +90^\circ)$ in both pitch and roll, covering the reachable workspace.

After resets, the quadrotor base spawns at the same initial position $(0.0, 0.0, 3.0)$~meters. However, the manipulator configuration is randomly sampled from $(-90^\circ, +90^\circ)$ for each joint, which encourages the quadrotor to learn to track poses from arbitrary arm configurations.

\subsubsection{Domain Randomization}

After every reset, we randomly sample a payload mass to be added to the end-effector from [-15g, 120g] and scale inertia accordingly. This results in a generalized policy that is able to perform pose tracking while carrying loads up to 140~g in the real world, shown in Section~\ref{result:real_world_paylaod}. 

We also observe noticeable differences in joint friction between the training and deployment environments, leading to variations in joint behavior under the same control inputs.
To address this, we first improve our simulation fidelity by finding friction and stiffness values that result in similar step responses from joint actuators across training, deployment simulators and the hardware. 
Once this gap is reduced, we apply domain randomization by scaling joint stiffness and friction to [0.75, 1.25] and [0.0, 1.5] times their default values, respectively.
The necessity of domain randomization is also analyzed through ablation studies in Section~\ref{sec:domain_randomization}.
\section{Real-World Experiments}
\label{chapter:real_results}
\begin{figure*}[t!]
    \includegraphics[width=1.0\textwidth]{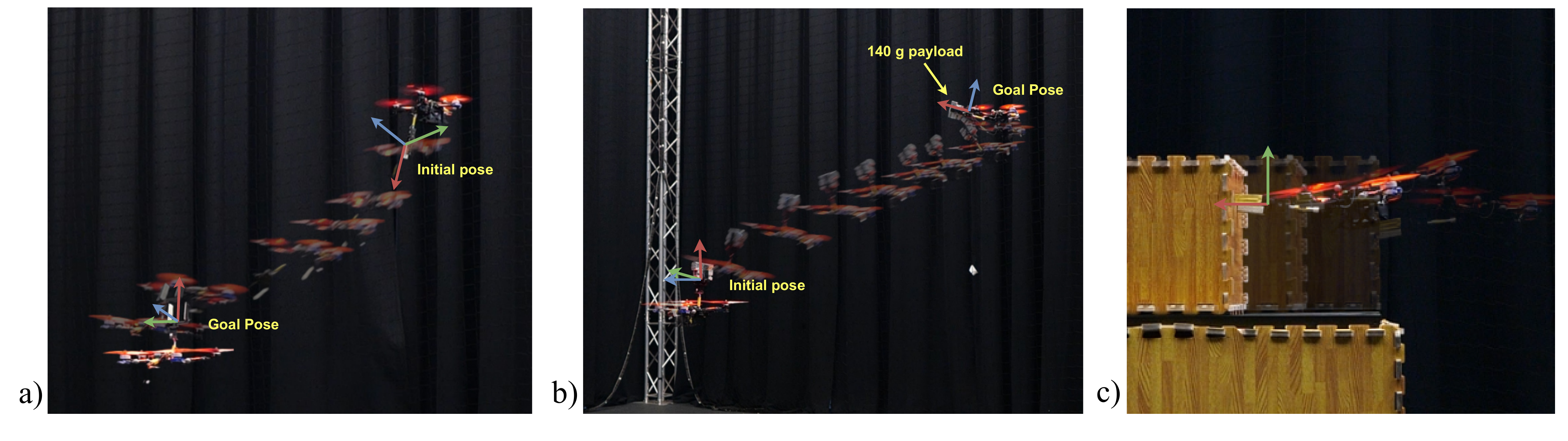}
    \captionof{figure}{a) End-effector pose control between two randomly sampled goal poses, b) Pose control with a 140~g payload held by the gripper, c) DSAM pushing a box weighing 590~g from right to left.}
    \label{fig:real_world}
\end{figure*}

\subsection{Experimental Setup}

The policy trained using the proposed method is deployed on our 2-DoF DSAM, with a total weight of 860g.
The moving part (arm plus gripper) weighs 75 g, actuated by two Dynamixel-XC330 motors~\cite{robotis_xc330m288} with embedded PID controllers following joint commands.
Our control architecture is developed upon Agilicious \cite{Foehn22science}, a modular software framework designed for agile quadrotor flight. 
The control stack entirely runs on the DSAM's onboard CPU (Raspberry Pi 5), including onboard inference using ONNX Runtime \cite{onnxruntime}. 
This results in an average inference time of \textbf{0.18 ms}.

Experiments are conducted in a 13 $\times$ 7~m facility equipped with a motion capture system that provides ground-truth pose measurements at 100~Hz, which is then fused with an EKF running onboard at 300~Hz to obtain the state estimates.
A safety controller that only controls the quadrotor base motion~\cite{sun2022comparative} is used for take-off and handles workspace violations by immediately bringing the system to hover state.
After take-off, control is switched from the safety controller to the learned policy that actuates the quadrotor to reach the commanded end-effector pose.

\subsection{End-Effector Pose Control}

We evaluate end-effector pose control on physical system by sending $10$ randomly sampled goal poses from a bounded workspace: $p_{x,goal} \in [-1, 1]$, $p_{y,goal} \in [-1, 1]$, $p_{z,goal} \in [1, 2]$ with roll and pitch angles from $-90^\circ$ to $90^\circ$ and yaw angle from $-120^\circ$ to $120^\circ$.
Each pose is maintained for 10 seconds, giving time for convergence, after which the next goal is sent. 
A pose is considered achieved, if the end-effector converges to the commanded pose without any intervention.
The snapshot of the experiment is given in Figure~\ref{fig:real_world}a.
The tracking performance is summarized in Table~\ref{tab:pose_tracking_error_stats_all} and presented in Figure~\ref{fig:sim2real_pose_tracking_5}.

\subsection{End-Effector Pose Control with Additional Load}
\label{result:real_world_paylaod}

To evaluate robustness under load we perform pose control while carrying payloads of 50 g and 140 g on 7 randomly generated goals. 
A snapshot of this experiment is presented in Figure~\ref{fig:real_world}b.

\begin{table}[h!]
\centering
\begin{adjustbox}{max width=\columnwidth}
\begin{tabular}{cccc}
\toprule
\textbf{Condition} & \textbf{Position Error (m)} & \textbf{Orientation Error (deg)} & \textbf{Success Rate} \\
\midrule
& mean \quad\quad std. & mean \quad\quad std. & (\# reached / total) \\
\midrule
0\,g   & 0.0536 \quad 0.0166 & 8.8078 ~\quad 7.1834  & 10/10 \\
50\,g  & 0.0995 \quad 0.0695 & 12.5020 \quad 3.0619 & 7/7 \\
140\,g & 0.0954 \quad 0.0505 & 15.7006 \quad 4.8312 & 7/7 \\
\bottomrule
\end{tabular}
\end{adjustbox}
\caption{Pose tracking error (mean and standard deviation) and success rate across payload conditions.}
\label{tab:pose_tracking_error_stats_all}
\end{table}

\begin{figure}[t!]
\centering
    \includegraphics[width=0.47\textwidth]{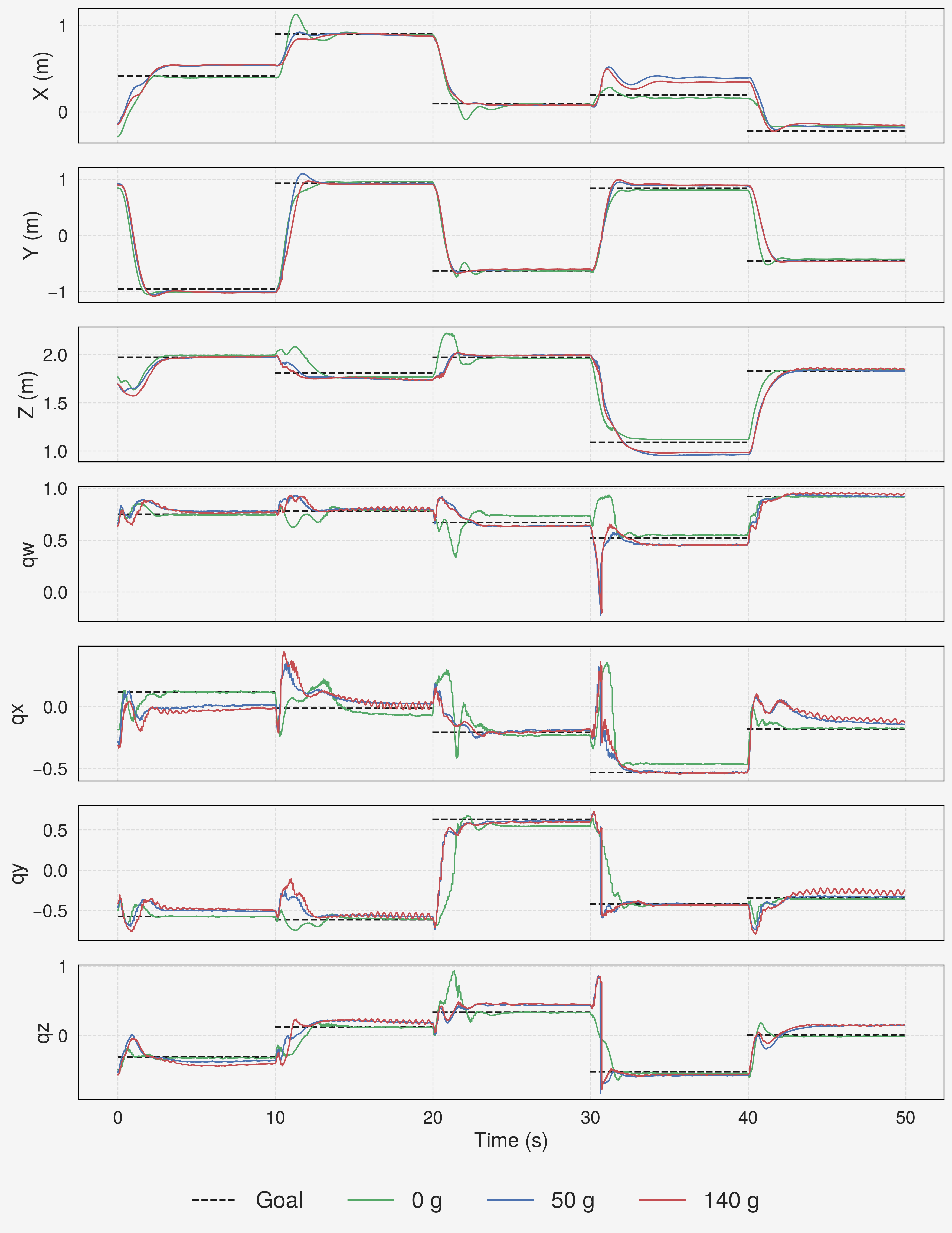}
    \captionof{figure}{Comparison of pose control performance with and without payload (we show only the first 5 setpoints for readability).}
    \label{fig:sim2real_pose_tracking_5}
\end{figure}

Our system maintains an average position error of ~10 cm for both 50 g and 140 g payloads. 
The orientation error increases by 3° with 140 g, which we attribute to the manipulator arm's higher sensitivity to load induced disturbances than the quadrotor base, visible from the quaternion component responses in Figure \ref{fig:sim2real_pose_tracking_5}. 
Position tracking errors are observed at extreme joint angles (visible from 30 - 40 s in Figure \ref{fig:sim2real_pose_tracking_5}). 

It is worth noting that with a 140~g payload, the moving part (arm plus load) accounts for 215~g, or 21.5\% of the total DSAM mass (DSAM plus load). 
Despite this, the policy maintains stable flight and demonstrates consistent convergence under load. 
The proposed controller also achieves agile transient responses with minimal overshoot, while preserving stability.
However, although domain randomization improves position tracking, we argue that it alone is still insufficient for accurate pose tracking under disturbances.
A greater pose error has been observed in comparison with the baseline.
An improved low-level controller capable of compensating for such disturbances may be required to further enhance tracking performance.

\subsection{Pushing an Object}

To evaluate robustness against contact-induced disturbances and the ability to exert forces on the environment, we task the policy with pushing a 590~g box along a straight line while maintaining a fixed end-effector orientation. We generate a straight-line path with constant end-effector orientation and provide the corresponding poses to the policy.
We then sample target poses along the path to ensure a constant end-effector velocity and feed them sequentially to the policy. Following these references, the DSAM approaches the box, establishes contact, and successfully pushes it.
During contact, the DSAM remains stable and continues to maintain end-effector orientation while adjusting the quadrotor attitude to exert a force, moving the box by 30 cm (Figure~\ref{fig:real_world}c). This result demonstrates that the learned whole-body controller can tolerate sudden interaction forces, despite being trained without explicit contact models or force feedback.

\subsection{Path following}
\begin{figure}[t!]
    \includegraphics[width=0.50\textwidth]{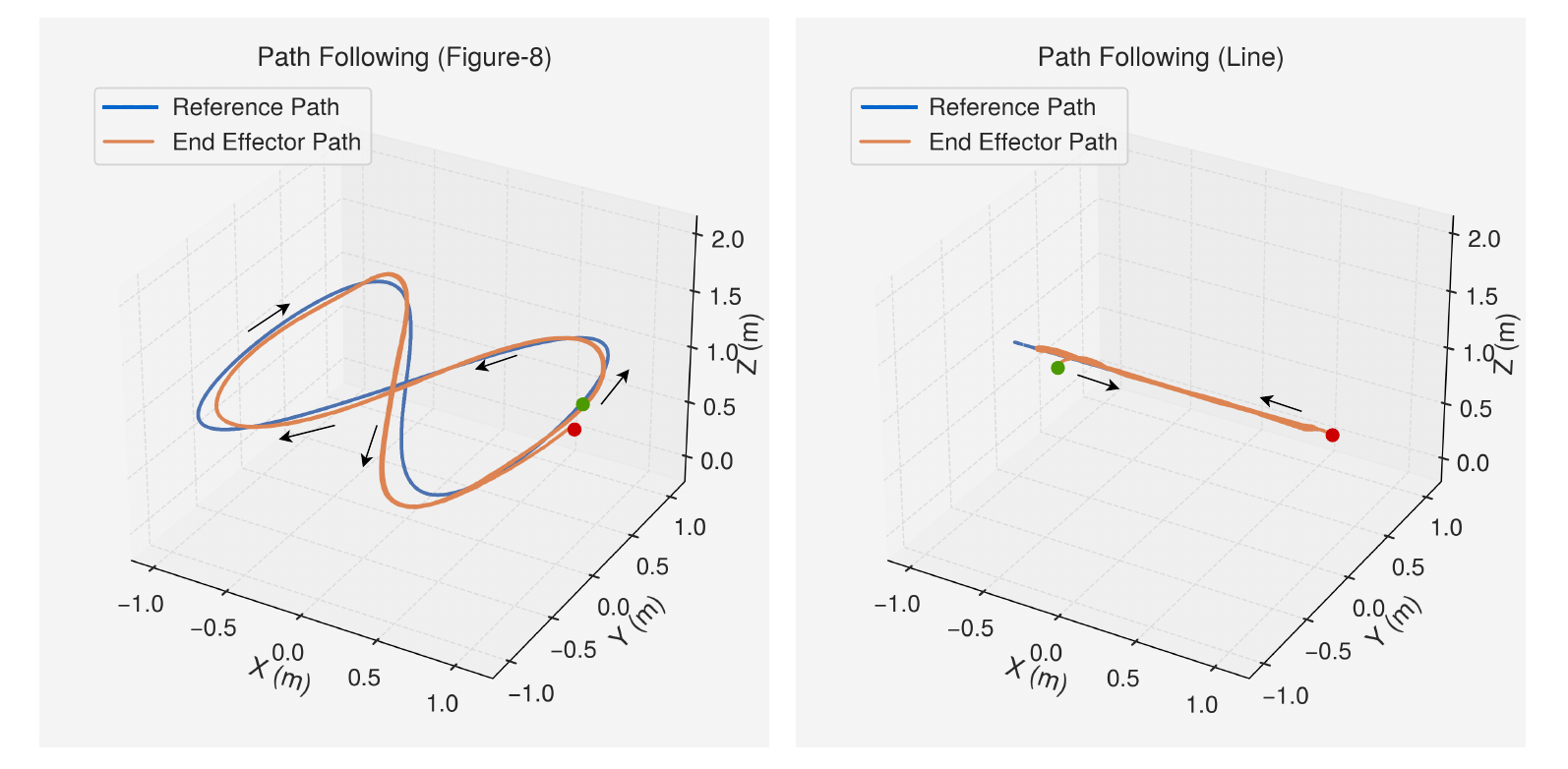}
    \captionof{figure}{3D trajectory of path followed by end-effector (starting point indicated by green dot)}
    \label{fig:3d_path}
\end{figure}

As an extension to pose control, we attempt path following without retraining. 
The end-effector is commanded to track a figure-eight path. 
The same as the pushing task, we achieve path following by sampling a sequence of target poses along the path and feeding them to the policy one at a time. 
The 3D trajectory during the path following task is presented in Figure~\ref{fig:3d_path}.

\begin{table}[h]
\centering
\begin{tabular}{lcc}
\toprule
\textbf{Path} & \textbf{Position RMSE (m)} & \textbf{Orientation RMSE (deg)} \\
\midrule
Figure-8      & 0.8167 & 14.3915 \\
Straight line & 0.1960 & 5.5607 \\
\bottomrule
\end{tabular}
\caption{RMSE of position and orientation errors for the two path following tasks.}
\label{tab:path_tracking_rmse}
\end{table}

As seen in Table~\ref{tab:path_tracking_rmse}, the RMSE for the figure-8 path is noticeably larger than for the straight line, reflecting the higher demands of multi-axis coordination compared to single-axis motion. The tracking accuracy is further limited because the policy is conditioned only on the immediate target pose and does not explicitly account for higher-order derivatives such as velocity, acceleration, or jerk. Consequently, the policy was not trained specifically for accurate path tracking but rather for pose regulation. Future work could address this by augmenting the observation space with future waypoints or target velocities, which have been shown to improve path-tracking performance in related learning-based control frameworks (e.g., ~\cite{garg2025dual}). Despite these limitations, the policy is able to follow simple paths while maintaining orientation, demonstrating whole-body coordination emerging from pose control alone.

\section{Ablation Studies}
\begin{figure*}[t!]
    \centering
    \includegraphics[width=1.0\linewidth]{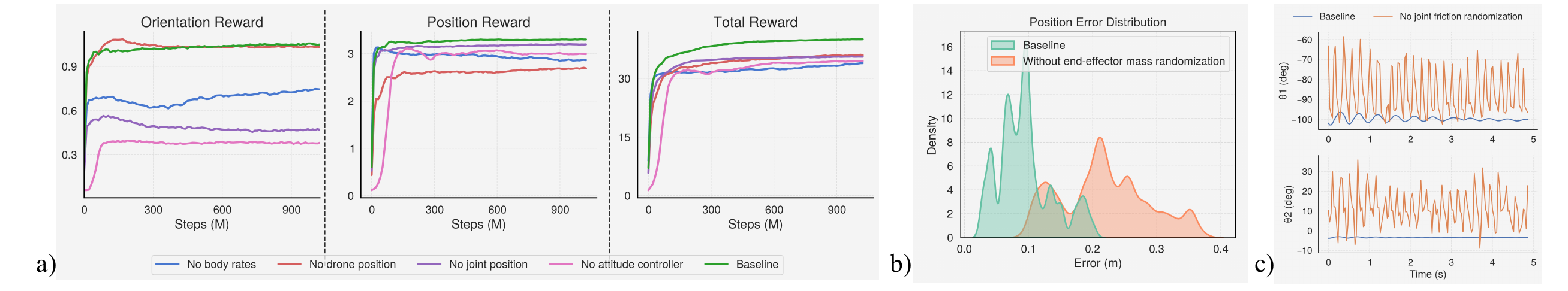}
    \caption{a) Learning curves with ablations in the training setup, b) Position error distribution on the end-effector pose control task carrying a 100~g load, without mass randomization, c) Oscillating joint position command from policy trained without friction randomization.
    }
    \label{fig:ablation}
\end{figure*}

To study the impact of design choices, we perform ablations on various components of our training setup, such as the observation space, low-level controller choice, domain randomization, and reward function.  We summarize the results in the learning curves, as well as position error distributions, presented in Figure~\ref{fig:ablation}.

\subsection{Observation Space Ablations}
\subsubsection{Joint positions} 
For a 2D-DSAM, through inverse kinematics, a unique mapping exists from the end-effector pose error and the goal position in the quadrotor frame (current observation) to the joint positions.
In other words, having joint positions leads to redundancies in the observation space.
However, despite being theoretically redundant to have joint positions in the observation, we notice that including the joint positions in the observation space leads to a significant increase in end-effector orientation reward.

\subsubsection{Goal positions in quadrotor frame}
In the observation space, we include the quadrotor position information by expressing the goal position in the quadrotor frame. 
Although these positions could, in principle, be derived from the goal pose in the end-effector frame and the joint angles, which have been included in the observation space, via inverse kinematics, omitting them leads to a substantial drop in position rewards.

A plausible explanation is that explicitly providing the joint positions and goal position in the quadrotor frame simplifies the learning problem by bypassing the need for the policy to infer this relationship implicitly, thereby improving training efficiency and tracking performance.

\subsubsection{Body rate of the quadrotor base}
We noticed that including body rate in the observation space is also critical for training for DSAM, which is particularly different from previous works on quadrotors~\cite{song2023reaching}, and AMs with a fully-actuated base~\cite{cuniato2023learning}.
We argue that including quadrotor angular rates in the observation significantly improves policy training, as they allow the policy to indirectly infer the effect of torque disturbances induced by the arm. This additional information helps the policy learn the coupled base–arm dynamics especially for DSAM with an underactuated unidirectional base, which is, on the other hand, less critical for omnidirectional platforms or simple quadrotors.

\subsection{Inner-Loop Controller Ablations} 
While incorporating an inner-loop is a common practice in RL-based control to achieve reliable sim-to-real transfer, determining the appropriate level of control abstraction is non-trivial. In particular, for the quadrotor base of DSAM, we replace the attitude controller with a simple proportional controller that maps the body-rate command to the desired body angular acceleration, which is then passed to the low-level INDI controller. Under this setup, we evaluate the traditional “collective-thrust plus body rate” (CTBR) output of the RL policy for quadrotor flight, following prior work~\cite{song2023reaching}. INDI is retained, as it plays a critical role in handling disturbances, as highlighted in previous studies~\cite{sun2022comparative,ferede2024end}. The training curves in Fig.~\ref{fig:ablation} show that our setup, where the attitude controller accepts reference acceleration (or equivalently attitude) from the RL policy, achieves consistently higher rewards in both orientation and position tracking. 

\subsection{Domain Randomization Ablations}
\label{sec:domain_randomization}

\subsubsection{End-effector mass} 

As shown in Fig.~\ref{fig:ablation}b, mass randomization leads to higher position accuracy while carrying additional load (100g), which can be attributed to the additional exploration induced by varying inertia. 
When exposed to different mass conditions, the agent is compelled to develop more robust control actions. 
In this sense, mass randomization serves as a form of controlled exploration, preventing the policy from overfitting to a single configuration and ultimately improving position-tracking performance.

\subsubsection{Joint friction: }

By removing joint friction randomization, we observe a substantial drop in orientation learning rewards.
Similar to the previous ablation, the setup with constant friction overfits to a narrow set of end-effector mass and joint stiffness values. When these properties are randomized with constant friction, a stick and slip motion is observed, resulting in unpredictable arm dynamics and reduced orientation learning. Without friction randomization, we observe oscillations in the joint actions generated by the policy (Figure \ref{fig:ablation}c).

\section{Conclusion}
\label{sec:conclusion}

We presented an RL-based control method for whole-body control of the DSAM, a minimal 2-DoF aerial manipulator capable of global 6-DoF end-effector pose control. 
By combining a PPO-based policy with INDI and PID low-level controllers and executing fully onboard with 0.18~ms of inference time, we achieved centimeter-level position accuracy and degree-level orientation accuracy in real-world experiments, while maintaining stable flight under disturbances, namely, carrying a payload on the end effector (over 16\% of the system mass) and pushing an object over 68\% of the system mass.

Our study suggests several future directions, including incorporating future waypoints or higher-order reference terms into the observation space to improve trajectory tracking, and extending the framework to contact-rich manipulation tasks. Overall, this work establishes the feasibility of learning-based whole-body control for DSAM, demonstrates its capability in real-world aerial manipulation, and lays the groundwork for increasingly accurate and versatile aerial manipulators.

\section{Acknowledgements}
This work was supported by the Dutch Research Council (NWO) through the Veni talent Programme (grant no. 20256, accurate aerial Manipulation).
We thank Jack Zeng for insightful discussions and for sharing low-level controller templates. We also thank Maurits Pfaff and Kseniia Khomenko for building the 2-DoF DSAM and for their support during experiments.

\printbibliography

\end{document}